# Multi-Pass, Multi-View Blended Learning for High-Fidelity Volumetric CT Synthesis from Chest X-Rays

Ozer Can Devecioglu[1], Serkan Kiranyaz[2], Rashid Mazhar[3], Tahir Hamid[3], Muhammad Chowdhury[2], Moncef Gabbouj[1]
[1]Department of Computing Sciences, Tampere University, Finland.
[2]Department of Electrical Engineering, Qatar University, Doha, Qatar.
[3]Hamad Medical Corporation, Doha, Qatar.

**Abstract— Reconstructing volumetric Computed Tomography (CT) from a single 2D chest radiograph (CXR) is an ill-posed inverse problem, further complicated by the scarcity of paired CXR/CT training data. Prior approaches address this by training on Digitally Reconstructed Radiographs (DRRs) which are synthetic projections derived from CT volumes. However, the domain gap between DRRs and real CXRs limits generalization, often resulting in coarse or anatomically inconsistent reconstructions when applied to clinical images. To address this challenging problem, this study introduces a Multi-Pass Multi-View Blended Learning for synthesizing high-fidelity volumetric CT directly from real chest X-ray (CXR) images. The proposed approach progressively decomposes the synthesis task into two distinct, complementary learning stages. Stage 1 is an unsupervised CXR-to-DRR Domain Adaptation, while Stage 2 includes three passes, namely, a) supervised DRR-to-CT Transformation, b) unsupervised Multi-View Slice Refinement, followed by c) Progressive Transfer Learning (PTL). With such a blended learning paradigm, the proposed approach mitigates the synthetic-to-real domain gap while enhancing both the structural integrity and anatomical detail of the final output. On the LIDC-IDRI dataset, where paired DRR–CT ground truth is available for quantitative evaluation, the proposed method improves upon prior methods by up to 14% in PSNR and 7.6% in SSIM. The framework successfully generates structurally consistent and anatomically realistic high-fidelity CT volumes from real CXRs, marking a significant advancement toward clinical viability of CT reconstruction from standard radiographic images.**

## I. INTRODUCTION

Computed Tomography (CT) is one of the most widely used imaging modalities in modern medicine. It provides detailed 3D representations of internal anatomical structures. Its ability to capture volumetric information makes it essential for tasks such as diagnosis, surgical planning, and treatment monitoring. However, acquiring CT scans comes at a cost. CT imaging exposes patients to relatively high doses of ionizing radiation compared to conventional X-ray imaging, raising concerns especially in repeated or long-term use. Ionizing radiation affects the human body by causing cellular damage at the atomic or molecular level; depending on the dose of exposure, this can result in cell damage or death (deterministic effects) or, more seriously, cancer transformation (stochastic effects). This is why "*as low as reasonably achievable*" (ALARA) is the guiding principle of diagnostic and interventional procedures using radiation. To put this in context: in the USA, people are exposed to average annual background radiation levels of about 3 mSv; exposure from a chest X-ray is about 0.1 mSv [1], and exposure from chest CT scans ranges from 1–1.5 mSv (low-dose CT for lung cancer screening) [2] to 7–18 mSv (high-resolution contrast chest CT) [3]-[4], depending on protocol and patient size. This means a chest CT can deliver radiation equivalent to 70–180 chest X-rays. In addition, CT scanners are expensive, not available usually in low-resource settings, and require more time and infrastructure to operate. Motivated by these challenges, recent research has explored reconstructing CT-like volumetric information directly from conventional chest X-ray (CXR) images.

Few studies have investigated deep-learning-based frameworks for this task. In one of the earliest methods, Ying et al. [6] proposed the X2CT-GAN framework, which integrates a 2D convolutional neural network (CNN), a 3D CNN, and generative adversarial networks (GANs). This architecture first extracts feature representations from 2D projections using 2D CNNs and subsequently perform volumetric reconstruction in an end-to-end manner through 3D CNN-based decoding. Beyond this study, due to the lack of directly paired CT/CXR datasets, most recent approaches rely only on Digitally Reconstructed Radiographs (DRRs) synthesized from the target CT to be reconstructed. Sun et al. [5] proposed INRR3T, a U-Net-based model that generates CT volumes from fewer than five planar simulated DRR projections. Zhang et al. [7] introduced a deep transformer model, XTransCT, which reconstructs volumetric CT data from simulated DRR images.

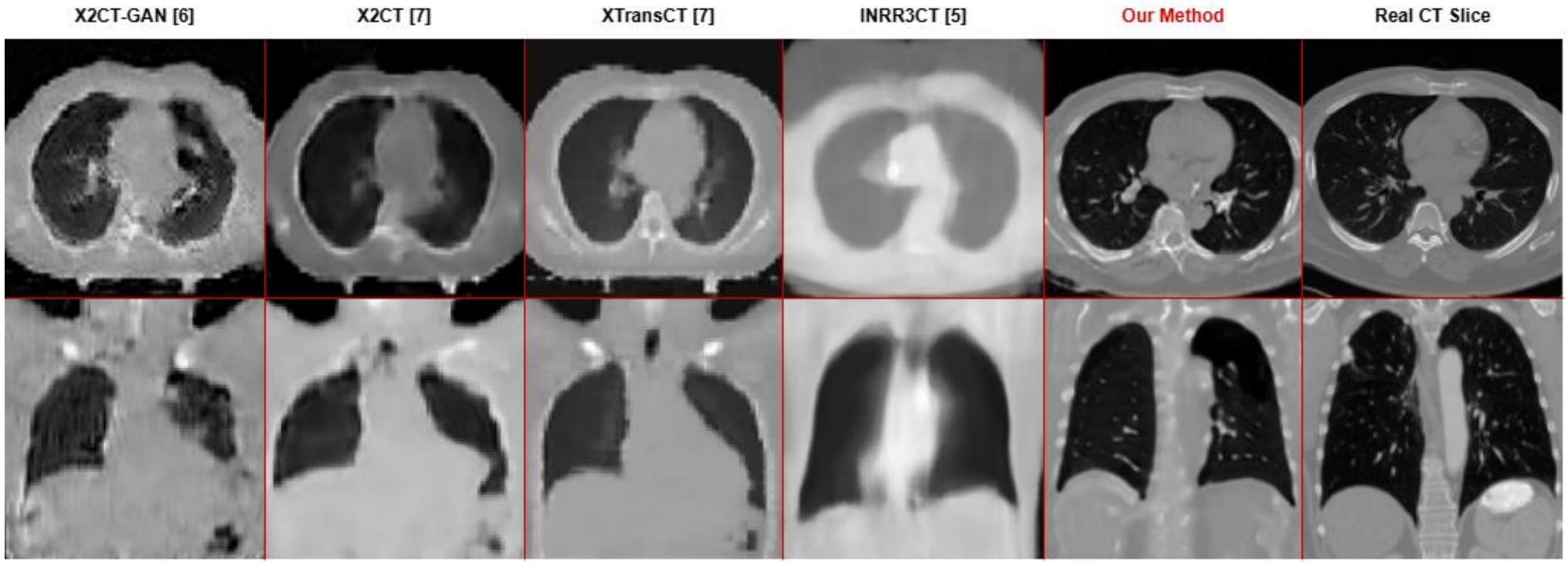

**Figure 1** Sample CT slices synthesized by the prior methods, INRR3CT [5], X2CT-GAN [6], X2CT [7], XTransCT [7] and the proposed approach. Slices are chosen from anatomically similar regions. **.**

While these methods make training easier and more controlled, extracting the source DRRs requires the target CT, which is obviously unavailable in real clinical use. Despite this increasing interest in 2D-to-3D reconstruction, DRR-to-volumetric-CT transformation remains fundamentally challenging. Mapping a limited number of 2D projections to synthesize a 3D volume is difficult, since the missing depth information must be inferred from a small number of 2D views. As shown in Figure 1, CT slices generated by recent models exhibit low-fidelity anatomical structure, often failing to preserve fine-grained spatial consistency and meaningful anatomical detail. Furthermore, the reconstructed slices resemble coarse, over-smoothed representations rather than realistic CT. This suggests that regardless of model depth or complexity, 2D-to-3D transformation cannot be effectively learned by a single model in a single supervised learning pass, even when trained on DRRs directly derived from the target CT. To address this challenge, we propose a Multi-Pass, Multi-View Blended Learning framework that decomposes CXR-to-CT synthesis into two stages, with the second stage applying three successive learning passes combining supervised and unsupervised learning. Since the primary objective is to transform *real* CXRs to volumetric CT, the two-stage framework is designed as follows. In Stage 1, the transformation of an input CXR to its corresponding DRR is learned in an unsupervised manner using an adversarial approach based on Operational GANs (Op-GANs) [24]. This transformation allows real CXR inputs to be converted into the DRR domain, enabling the proposed approach to operate on real clinical data rather than only on synthetic DRRs.

Stage 2 performs the actual CT reconstruction over the synthesized DRR through a multi-pass process involving three successive passes. In the first pass, a DRR-to-CT mapping is learned in a supervised manner using synthetic DRR–CT pairs, producing an initial CT reconstruction that captures the correct anatomical layout but lacks fine detail. In the second pass, this initial volume is refined using a multi-view slice-based strategy: the volume is decomposed into axial, sagittal, and coronal views, and each view is independently enhanced through unsupervised adversarial learning. . Because this pass learns the target CT domain directly rather than a paired transformation, we will demonstrate that it substantially improves the structural consistency, the anatomical details, and the overall visual quality. In the third and final pass, Progressive Transfer Learning (PTL) [25] is applied to further boost synthesis performance. PTL restarts the optimization process at intermediate stages, allowing the model to continue improving beyond standard convergence, helping it escape performance plateaus and produce sharper, more realistic CT reconstructions. As shown in Figure 1 (fifth column), this two-stage, multi-pass design allows the proposed method to preserve key anatomical structures — including the lung regions and central thoracic anatomy — while maintaining accurate structural boundaries with improved spatial coherence.

The novel and significant contributions of this study can be summarized as follows:

- We propose Multi-Pass, Multi-View Blended Learning, within a two-stage framework for synthesizing high-fidelity and realistic volumetric CT views from real CXR images.
- We design a multi-view slice-based refinement method, in which volumetric CT outputs are decomposed into axial, coronal, and sagittal views and independently enhanced using adversarial learning to improve cross-view structural consistency.
- We incorporate Progressive Transfer Learning (PTL) to iteratively refine reconstruction quality beyond convergence, enabling the model to escape local performance plateaus and enhance anatomical realism.
- Finally, compared to recent DRR-to-CT transformation methods, an extensive set of qualitative and quantitative evaluations demonstrate that the proposed approach produces structurally consistent, anatomically realistic, and clinically interpretable 3D CT synthesis.

The rest of the article is organized as follows. Section II describes the proposed methodology in detail. Section III presents experimental setup, evaluation metrics, qualitative and quantitative results with comparative evaluations. Section IV provides concluding remarks and suggests topics for future research.

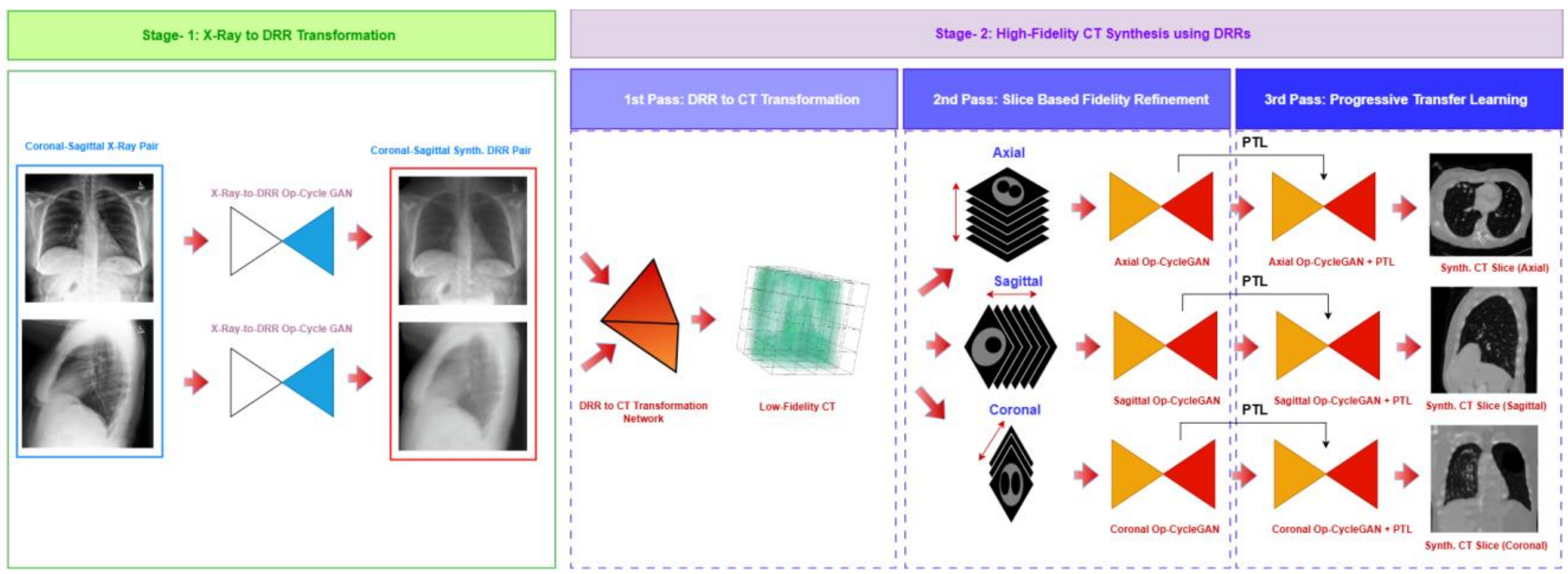

**Figure 2** Block diagram of the proposed approach for CT Synthesis over real CXR images.

## II. Methodology

As shown in Figure 2, the proposed Multi-Pass, Multi-View Blended Learning framework is implemented in two consecutive stages. The first stage performs unsupervised domain translation from real CXR images to the DRR domain, enabling the reconstruction network to process real clinical data. In the second stage, over the DRRs synthesized in the first stage, high-fidelity volumetric CT is synthesized through three successive learning passes. To this end, Self-Organized Operational Neural Networks (Self-ONNs) [8]-[12] are incorporated into the domain translation and refinement stages to enhance representation and learning capability. During the multi-pass blended learning, Progressive Transfer Learning (PTL) is employed during the final pass to boost the unsupervised learning performance. Self-ONNs and PTL are summarized in the following two subsections; the two-stage CT synthesis pipeline is then detailed in the final two subsections.

### *A. Self-Organized Operational Neural Networks*

In contrast to conventional CNNs, which rely on fixed convolutional operations, Self-ONNs [9] employ generative neurons equipped with nodal operators that enable adaptable nonlinear transformations. These transformations can be approximated through a Taylor series expansion around a point near the origin.

$$\psi(x) = \sum_{n=0}^{\infty} \frac{\psi^{(n)}(0)}{n!} x^n \tag{1}$$

This formulation allows for the approximation of any arbitrary function ψ near zero. When activation functions such as the hyperbolic tangent (tanh) constrain the input feature maps of the neurons within a region close to zero, the expression in Equation (1) can be leveraged to construct a composite nodal operator. Denoting $\frac{\psi^{(n)}(0)}{n!}$ as wn, the nodal operator can be expressed as:

$$\psi(w, y) = w_0 + w_1 y \quad + w_2 y^2 + \cdots + w_Q y^Q \tag{2}$$

Here, $w_0$ represents the bias term, while $w_1$ through $w_Q$ correspond to the Maclaurin series coefficients, all learned during backpropagation. For a more detailed explanation of the theoretical foundation, forward propagation mechanism and applications of Self-ONNs, readers are referred to [8]-[23].

### *B. Progressive Transfer Learning (PTL)*

This section introduces Progressive Transfer Learning (PTL), a model-agnostic multi-pass transformation learning strategy designed to progressively improve the quality of generated images. Unlike conventional single-pass approaches, which attempt to learn a direct mapping in one step, PTL decomposes the transformation process into a sequence of incremental refinements applied over multiple passes. Directly learning a complex transformation is challenging due to the presence of multiple artifacts and the ill-posed nature of the problem; a single model may fail to resolve all inconsistencies simultaneously. PTL addresses this by iteratively refining the output through multiple passes, with each pass improving the residual errors of the previous one.

Let $X^{(0)}$ denote the initial input. At the first pass, a transformation model $M^{(1)}$ is trained to map $X^{(0)}$ to to an improved representation. The best-performing model is selected according to a task-specific objective loss function ($L^{(1)}$),

$$M^{(1)} = \arg\min_{M} (L^{(1)}, X^{(0)}) \tag{3}$$

This approach enables the evaluation of the realism of the outputs for subsequent regression. The refined output is then obtained as,

$$X^{(1)} = M^{(1)}\left(X^{(0)}\right) \tag{4}$$

At each subsequent pass, $i + 1$, a new model $M^{(i+1)}$is initialized using the weights of the previous best-performing model,$M^{(i+1)}$, enabling knowledge transfer across passes. The optimal model at each pass is selected as:

$$M^{(i+1)} = \arg\min_{M} (L^{(i+1)}, X^{(i)}) \tag{5}$$

and applied to further improve the output:

$$X^{(i+1)} = M^{(i+1)}\left(X^{(i)}\right) \tag{6}$$

This iterative process allows the model to progressively enhance structural consistency and fine-grained detail while reducing residual artifacts. By reusing learned representations across passes, PTL also mitigates catastrophic forgetting and stabilizes training. The final output, $X^{final}$, after $n$ passes is obtained through sequential transformations:

$$X^{final} = M_n\left[M_{n-1}\left[\dots M_1\left[X^{(0)}\right]\right]\right] \tag{7}$$

PTL converts a challenging one-step transformation problem into a sequence of simpler sub-problems, improving training stability and effectiveness. For a more detailed explanation of the theoretical foundation of PTL, readers are referred to [25].

### C. ***Stage-1***: *CXR-to-DRR Domain Transformation*

Since the CT reconstruction models in Stage 2 are trained on paired DRR–CT data, they cannot be applied directly to real CXR images due to the domain discrepancy between DRRs and CXRs. As illustrated in Figure 3, models trained solely on DRR images fail to generalize to real CXRs, making an intermediate domain adaptation step essential. Moreover, paired CT data exists only for DRRs generated from actual CT volumes, not for real CXRs. To enable the use of the DRR-trained reconstruction models on real clinical CXRs, Stage 1 performs an unsupervised CXR-to-DRR transformation using an Op-CycleGAN [24]

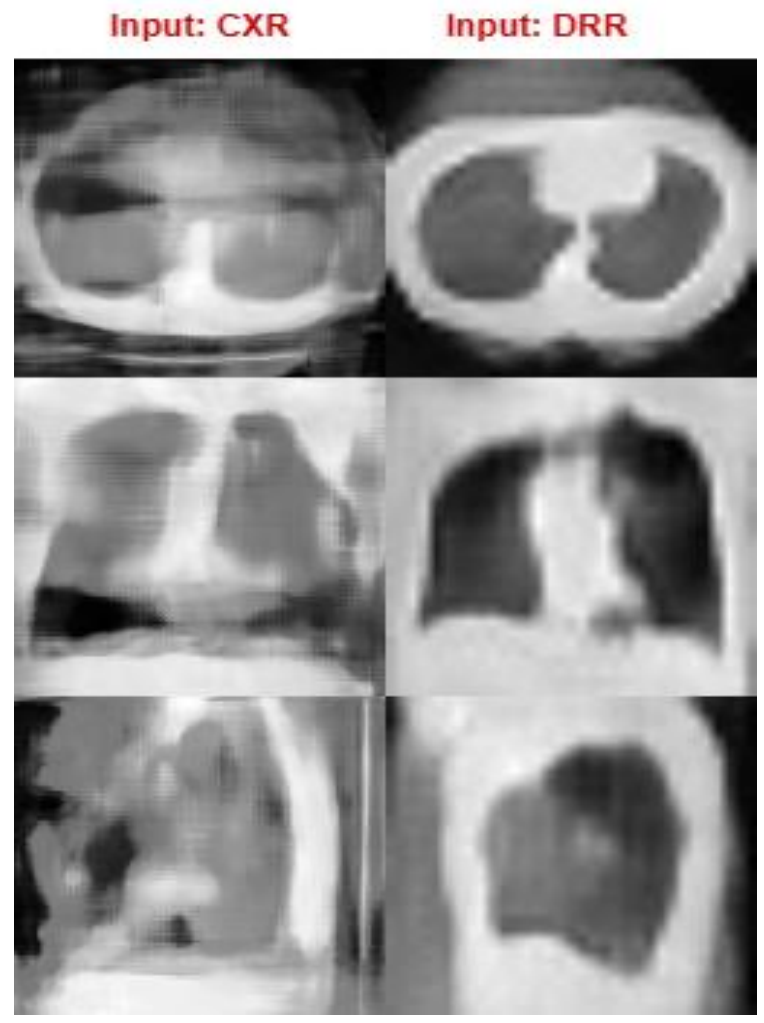

**Figure 3** Transformation failures of DRR-trained models on real CXR images due to domain discrepancy.

Real CXR images from the CheXpert dataset [26] are used to train the domain translation model. All CXR images are resized to 128×128 pixels prior to training. For the CXR-to-DRR translation stage, 818 frontal and lateral X-ray pairs are randomly selected from CheXpert. The Op-CycleGAN architecture is trained here, with 818 DRR pairs generated from the LIDC-IDRI dataset [27] used as the target domain. Since no one-to-one correspondence exists between the two datasets, the model learns the mapping between the X-ray and DRR distributions without paired supervision. The remaining CheXpert images are reserved for testing and evaluation.

### D. ***Stage-2***: *High-fidelity CT Synthesis over DRR Images*

Stage 2 generates individual high-fidelity CT views (axial, sagittal, and coronal) through three successive passes. In the first pass, a structurally consistent but low-fidelity baseline volumetric CT is synthesized from the two 2D DRR images using supervised learning over paired DRR–CT data, preserving the major anatomical structures. In the second pass, the low-fidelity volumetric CT is refined using an unsupervised (unpaired) learning strategy via an Op-CycleGAN , recovering anatomical details and patterns missed in the first pass. In the third and final pass, PTL is applied to iteratively enhance synthesis quality, further improving anatomical fidelity and yielding the final high-fidelity volumetric CT view. The full pipeline is illustrated on the right side of Figure 2.

#### D.1. **First pass** — DRR-to-CT baseline transformation

The network architecture used for the DRR-to-CT transformation in the first pass is shown in Figure 4. The DRR images, $X^{DRR}_{\text{Coronal}}$ and $X^{DRR}_{\text{Sagittal}}$, are unit-normalized to [0, 1]. Each input is processed independently through two identical 2D encoder networks to extract high-level feature representations:

$$Y_{\text{Coronal|Sagittal}} = E_{\text{Coronal|Sagittal}}(X^{DRR}_{\text{Coronal|Sagittal}}) \tag{8}$$

The resulting feature maps are then concatenated along the channel dimension to form a unified representation:

$$Y_{\text{Concat}} = H(Concat(Y_{\text{Coronal}}, Y_{\text{Sagittal}})) \tag{9}$$

where $H(\cdot)$ denotes the projection operation implemented via a $1 \times 1$ convolution followed by reshaping into a 3D tensor with predefined depth bins. The projected volume is further processed using a 3D convolutional refinement network based on a U-Net architecture [30], which enhances spatial coherence and aggregates contextual information across the volume to produce a coarse CT reconstruction:

$$V^{LQ}_{CT} = Decoder_{\text{3D}}(Y_{\text{Concat}}) \tag{10}$$

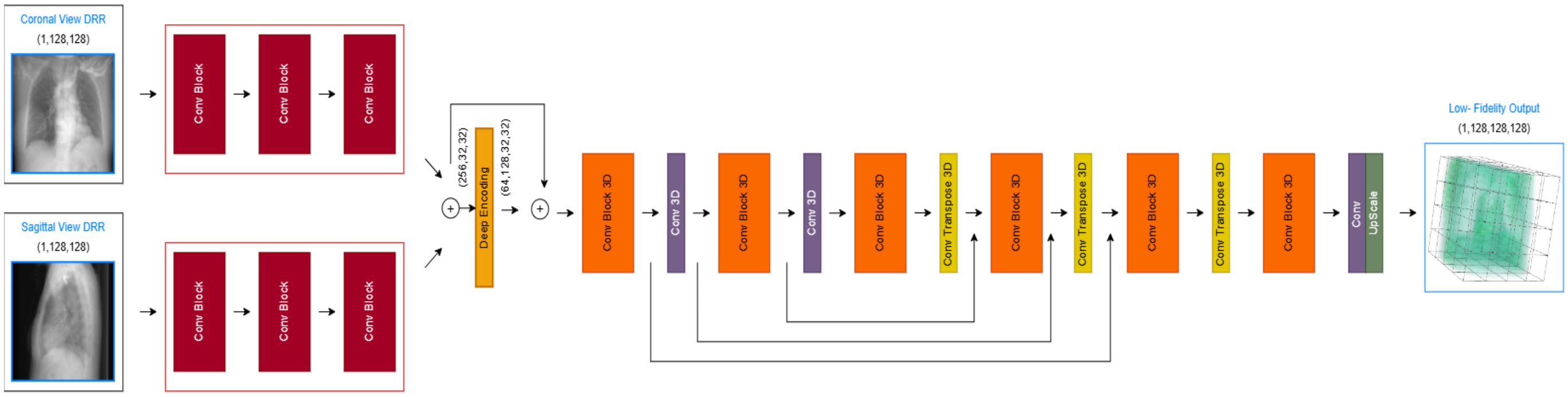

**Figure 4** Network architecture used for the DRR-to-CT transformation in the first pass. The 2D convolutional block (Conv Block 2D) consists of two sequential Conv2D–Batch Normalization–ReLU layers, while the 3D convolutional block (Conv Block 3D) consists of two sequential Conv3D–Batch Normalization–ReLU layers.

The output of this pass, $V_{CT}^{LQ}$, is a low-fidelity CT volume that preserves the overall anatomical structure but lacks fine-grained detail and texture. The model is trained using a combination of voxel-wise reconstruction loss and structural similarity constraints. First, an $L1$ loss minimizes the absolute difference between the predicted volume and the ground truth volumes:

$$\mathcal{L}_{L1} = \| V_{CT}^{LQ} - V_{CT}^{gt} \|_1 \quad (11)$$

In addition, a multi-view Structural Similarity Index Measure (SSIM) loss is incorporated to better preserve anatomical structures across different orientations, computed on axial, coronal, and sagittal slices, and averaged:

$$\mathcal{L}_{SSIM} = \frac{1}{3} \| (SSIM_{axial} + SSIM_{coronal} + SSIM_{sagittal}) \|_1 \quad (12)$$

The final first-pass loss is a weighted combination of the two terms:

$$\mathcal{L}_{total} = \lambda_1 \mathcal{L}_{L1} + \lambda_2 \mathcal{L}_{SSIM} \quad (13)$$

This formulation encourages reconstructions that are voxel-wise accurate and structurally consistent across anatomical planes.

### D.2. ***Second pass*** *— multi-view slice refinement*

After generating initial low-fidelity CT volumes in the first pass, this pass enhances anatomical detail and overall reconstruction quality with unsupervised learning. We employ an Op-CycleGAN that learns bidirectional mappings between the low-quality (LQ) and high-quality (HQ) CT domains in an unpaired (unsupervised) manner, restoring fine anatomical structures and details while preserving global structural consistency. As shown in Figure 5 the Op-CycleGAN consists of two operational generators, $OG_{LQ\to HQ}$ and $OG_{HQ\to LQ}$, and two operational discriminators, $OD_{HQ}$ and $OD_{LQ}$. The generators perform bidirectional translation between the low- and high-quality domains, while the discriminators learn to distinguish synthesized images from real samples within each domain. To preserve anatomical structure during translation, a cycle-consistency loss is applied:

$$\begin{aligned} L_{cyc}&(OG_{LQ\to HQ}, OG_{HQ\to LQ}, X_{LQ}, X_{HQ}) \\ &= \frac{1}{m}\sum_{i=1}^{m}\left[|OG_{HQ\to LQ}\left(OG_{LQ\to HQ}\left(X_{LQ}(i)\right)\right) - X_{LQ}(i)|_1\right] \\ &+ \frac{1}{m}\sum_{i=1}^{m}\left[|OG_{LQ\to HQ}\left(OG_{HQ\to LQ}\left(X_{HQ}(i)\right)\right) - X_{HQ}(i)|_1\right] \end{aligned} \quad (14)$$

. Adversarial losses are applied to both mappings:

$$\begin{aligned} L_{adv1}(OG_{LQ\to HQ}, OD_{HQ}, X_{LQ}) &= \frac{1}{m}\sum_{i=1}^{m}\left(1 - OD_{HQ}\left(OG_{LQ\to HQ}\left(X_{LQ}(i)\right)\right)^2\right) \\ L_{adv2}(OG_{HQ\to LQ}, OD_{LQ}, X_{HQ}) &= \frac{1}{m}\sum_{i=1}^{m}\left(1 - OD_{LQ}\left(OG_{HQ\to LQ}\left(X_{HQ}(i)\right)\right)^2\right) \end{aligned} \quad (15)$$

These terms encourage visually realistic outputs in each domain. An identity loss is also included to prevent unnecessary modifications:

$$\begin{aligned} L_{ide}&(OG_{LQ\to HQ}, OG_{HQ\to LQ}, X_{LQ}, X_{HQ}) \\ &= \frac{1}{m}\sum_{i=1}^{m}\left[|OG_{LQ\to HQ}\left(X_{HQ}(i)\right) - X_{HQ}F(i)|_1\right] \\ &+ \frac{1}{m}\sum_{i=1}^{m}\left[|OG_{HQ\to LQ}\left(X_{LQ}(i)\right) - X_{LQ}(i)|_1\right] \end{aligned} \quad (16)$$

The final Op-CycleGAN objective combines all terms:

$$L_{total} = L_{adv1} + L_{adv2} + \lambda L_{cyc} + \beta L_{ide} \quad (17)$$

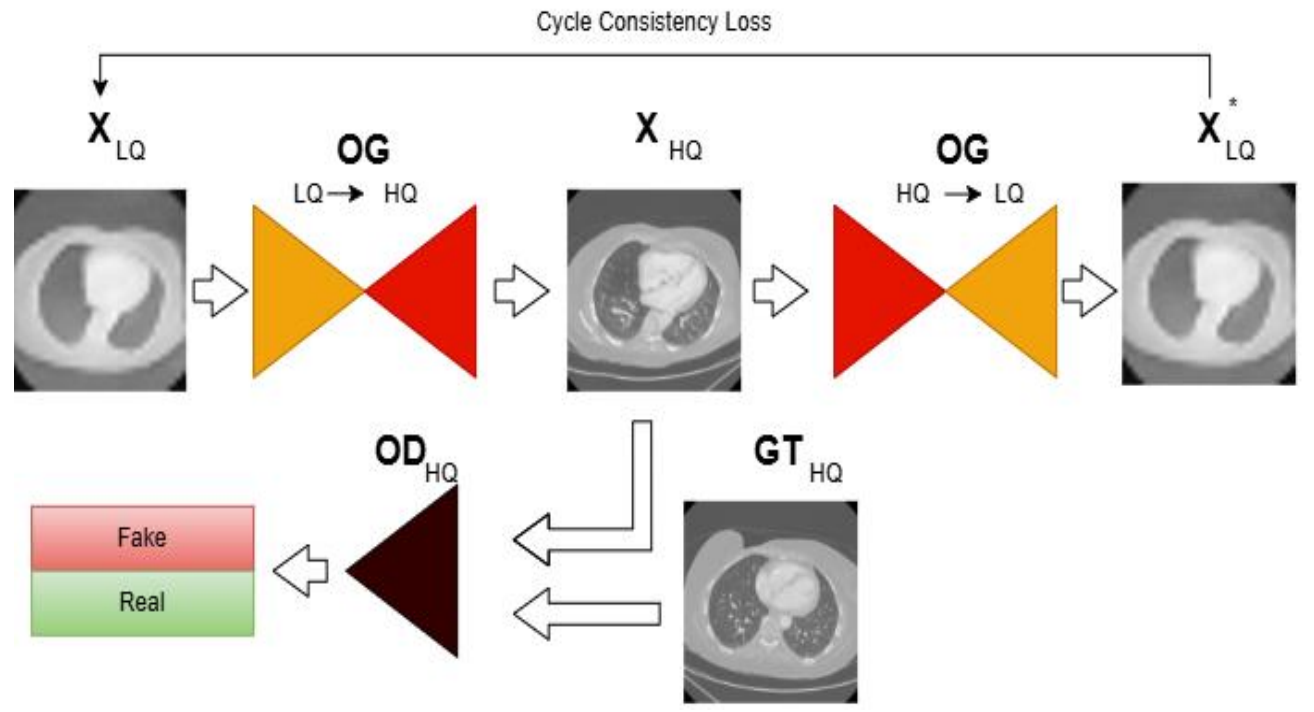

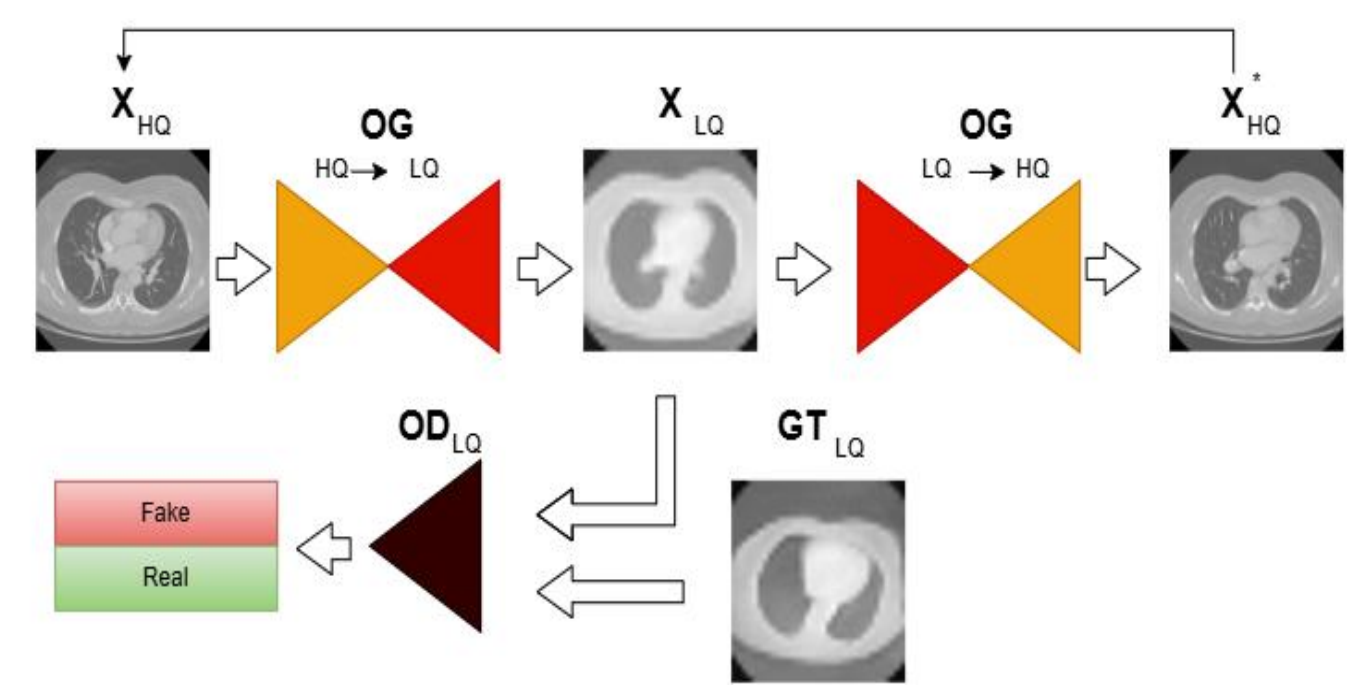

**Figure 5** General Framework of Op-CycleGAN for CT slice refinement. Starting from the left side, $\boldsymbol{OG_{LQ\to HQ}}$ converts a low-quality CT slice image into a high-quality counterpart. $\boldsymbol{OD_{HQ}}$ assesses whether the generated image is real or fake, imposing adversarial loss against real high-quality images. $\boldsymbol{OG_{HQ\to LQ}}$then reconstructs the original low-quality image, maintaining cycle consistency. The reverse process occurs on the right panel, where. $\boldsymbol{OG_{HQ\to LQ}}$ maps a high-quality image to a low-quality version, and $\boldsymbol{OD_{HQ}}$ distinguishes real low-quality images from generated ones.

#### *D.3* ***Third pass*** *— Progressive Transfer Learning*

To the final pass further enhances the CT views produced in the second pass. The same Op-CycleGAN architecture is trained independently for each anatomical view, with each model initialized using PTL as described in Section II.B. As a result, each CT view (axial, coronal, sagittal) is individually refined from its corresponding synthesized DRR input.

## III. EXPERIMENTAL RESULTS

This section presents the experimental evaluation of the proposed framework. First, the datasets, implementation details, and evaluation metrics are introduced. Next, quantitative and qualitative results are presented for CT reconstruction from DRR images, followed by the qualitative results using real X-ray images. Finally, the computational complexity of the proposed framework is analyzed. Although no formal ablation study is provided, intermediate outputs of each stage and after each pass ($2^{nd}$ and $3^{rd}$ passes) are provided to show the improvements brought by the respective module.

### *A. LIDC-IDRI and CheXpert Datasets*

The proposed framework is evaluated using two publicly available datasets, each serving a distinct role in the pipeline. The LIDC-IDRI Dataset [27] contains 1,018 thoracic CT scans acquired for lung cancer screening and diagnosis, 157 of which belong to subjects diagnosed as cancer. All CT volumes are resampled to a fixed resolution of 128×128×128, with intensity values normalized between 0-1. For each CT scan, frontal and lateral Digitally Reconstructed Radiographs (DRRs) are generated and used as inputs to the Stage 2 reconstruction network. Of the 1018 scans, 818 CT volumes (and their corresponding DRRs) are used for training and validation (The training dataset was divided such that 10% of the training data was used for validation.), while the remaining 200 volumes are reserved for testing. Because paired DRR–CT ground truth is available for this dataset, all quantitative results (PSNR, SSIM) reported in this section are computed on LIDC-IDRI.

The CheXpert dataset [26] contains 224,316 CXR images, of which 65,240 ae labeled with health anomalies including pneumonia; 59.4 % of the images belong to male subjects, with the reminder belonging to female subjects. For Stage 1 training (CXR-to-DRR domain adaptation), 818 frontal and lateral X-ray pairs are randomly selected from CheXpert, matching the number of DRR pairs used from LIDC-IDRI. For qualitative evaluation on real clinical CXRs (Section III.E), we additionally selected 2,000 CXRs, equally divided between healthy men and women. Since no paired CT ground truth exists for CheXpert subjects, all CheXpert-based results reported in this section are qualitative.

### *B. Experimental Setup*

The network architecture used in the first pass (DRR-to-CT baseline transformation) is illustrated in Figure 4. The proposed model consists of two parallel 2D encoders, one for each DRR view, followed by feature concatenation, a 2D-to-3D projection module, and a 3D U-Net-based refinement network. Each 2D encoder contains three convolutional blocks with 32, 64, and 128 output channels, respectively, with two max-pooling operations between the blocks. The fused 256-channel feature representation is projected into a 3D feature volume using a convolution and reshaping operation, followed by a 3D convolutional layer with 64 output channels. The refinement network employs a U-Net-like encoder–decoder structure with three skip connections and three 3D transposed convolutional layers. The final convolution produces a single-channel volumetric output, which is subsequently resized using trilinear interpolation. Training is performed using the AdamW optimizer with an initial learning rate of $1\times10^{-5}$, together with a cosine annealing learning-rate scheduler. The batch size is set to 2 for training and 1 for validation. The model is trained for up to 2000 epochs.

For the second pass, the Op-CycleGAN uses an operational ResNet-based generator consisting of 9 residual blocks, with channel outputs starting at 16 kernels in the first layer and no dropout. The discriminator is composed of six operational layers. Training is performed for up to 3,500 epochs, with the learning rate initially set to $2\times10^{-4}$ for the first 100 epochs, followed by a linear decay to $1\times10^{-5}$ over the next 100 epochs, for both generator and discriminator.

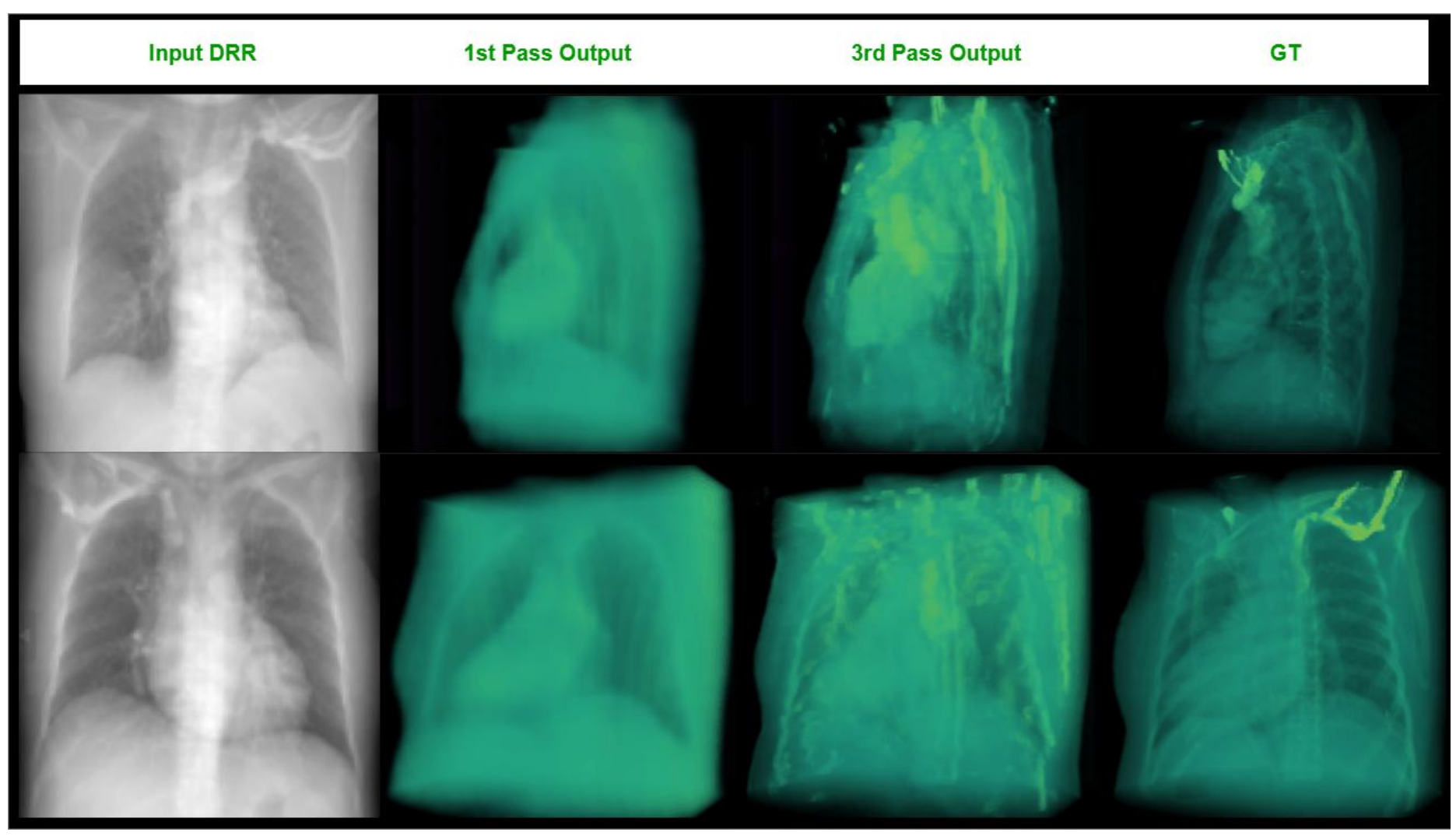

**Figure 6** 3D Sample CT synthesis results at different passes of the proposed approach.

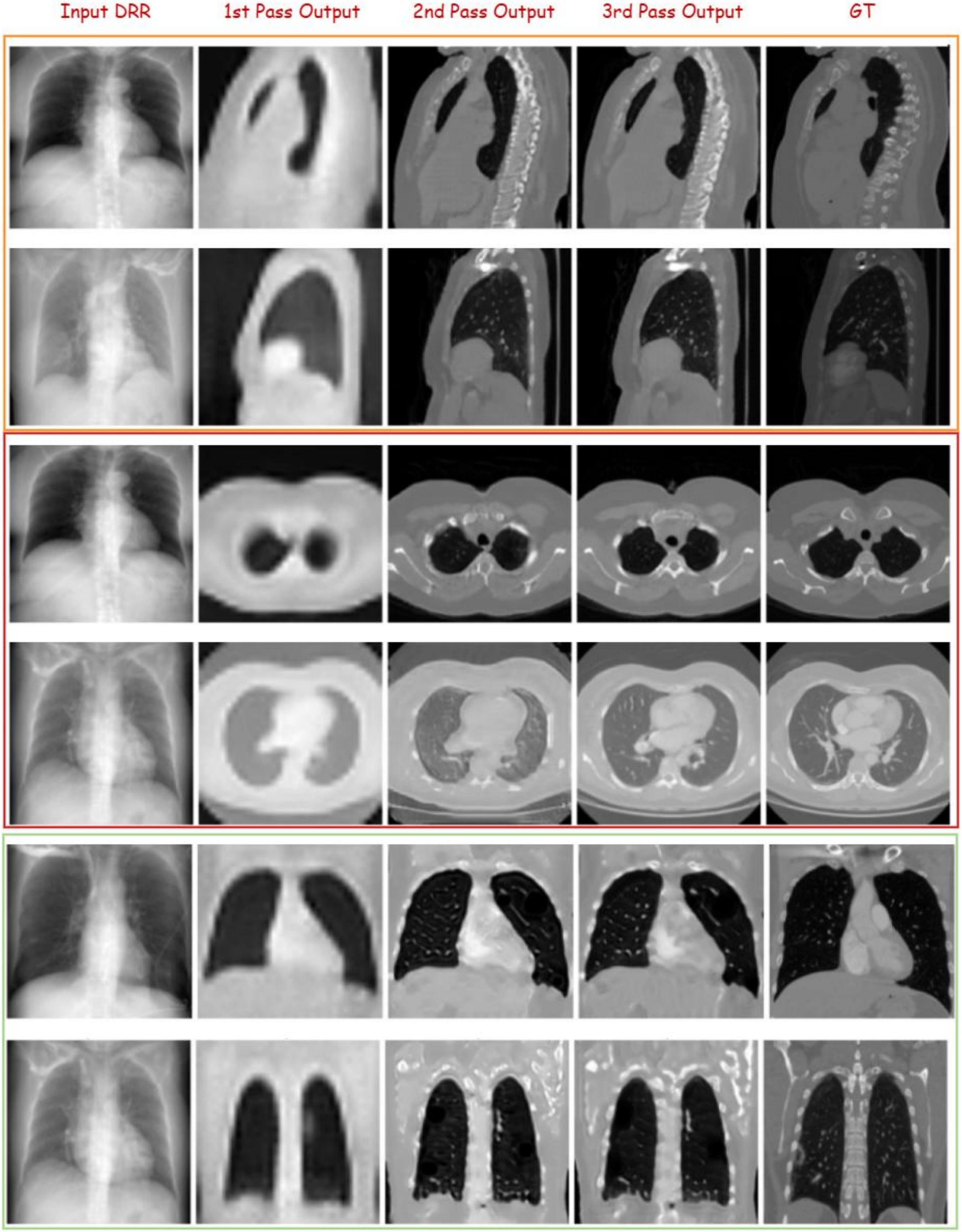

**Figure 7** Sample CT slice synthesis results at different passes of the proposed approach.

**Table 1** Quantitative comparison of CT reconstruction results at different stages of the proposed framework

| | Axial | | Coronal | | Sagittal | |
|---|---|---|---|---|---|---|
| | PSNR (dB) | SSIM | PSNR (dB) | SSIM | PSNR (dB) | SSIM |
| **IRR3CT [5]** | 14.41 | 0.5432 | 13.62 | 0.4693 | 13.06 | 0.4649 |
| **First Stage Output** | 14.29 | **0.5513** | 13.78 | 0.4843 | 12.87 | 0.4683 |
| **Op-CycleGAN** | 15.22 | 0.5114 | 15.18 | 0.4820 | 14.73 | 0.4959 |
| **Op-CycleGAN + PTL** | **15.49** | 0.5254 | **15.39** | **0.4986** | **14.93** | **0.5005** |

### *C. CT Synthesis over DRRs*

This section presents qualitative and quantitative evaluations of CT synthesis from DRR images (LIDC-IDRI). Each pass is evaluated independently to demonstrate its contribution to overall reconstruction performance, using PSNR and SSIM alongside qualitative comparisons. Figure 6 presents sample 3D CT synthesis results across passes. From left to right, the columns show: (1) the input DRR images, (2)–(3) the first- and third-pass outputs, and (4) the ground-truth (GT) volumetric CT. The same trend holds in 3D, with the final synthesized volume showing markedly enhanced anatomical detail in structures such as bone, heart, and other soft tissue. Additional 3D results can be explored interactively via the study's GitHub page [28].

Figure 7 shows two sample CT slice synthesis results illustrating the improvement achieved by each pass. From left to right, the columns show: (1) the input DRR images, (2)–(4) the three-pass CT synthesis outputs, and (5) the corresponding ground-truth CT slices. Sagittal, axial, and coronal slices are shown in rows to evaluate the three-dimensional consistency of the synthesized volumes. The results indicate that the first pass generates the main anatomical structure but with low-fidelity synthesis, blurred boundaries, and little to no fine detail. In the second pass, Op-CycleGAN refinement recovers much of the missing anatomical structure and detail, yielding markedly improved contrast and reduced blurring. In the third and final pass, PTL further sharpens structural boundaries and textures, improving the preservation of fine anatomical detail. Due to space limitations, further slice-based results are provided in the Appendix (e.g., see: Figures 9–11).

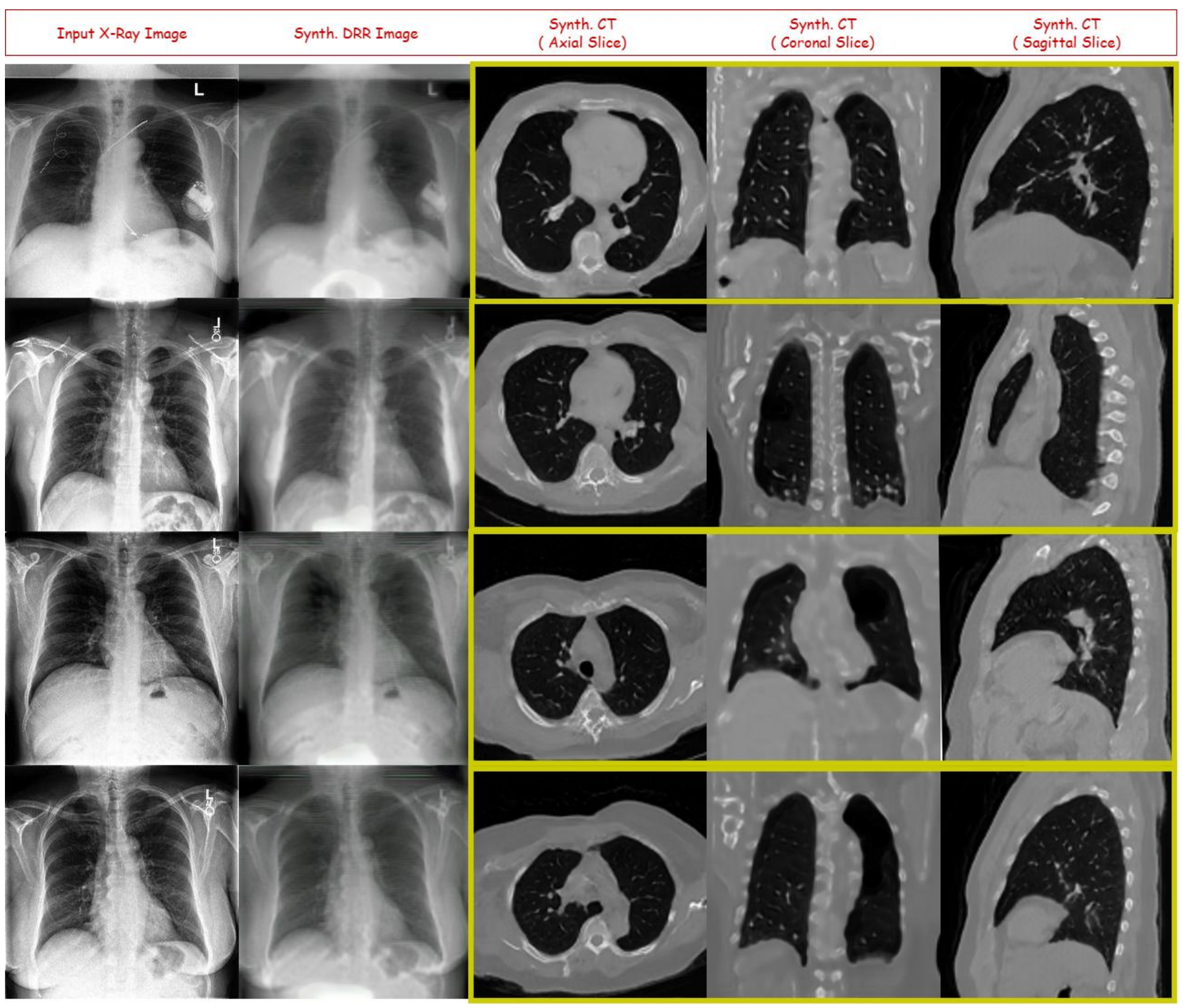

**Figure 8** Slice based CT reconstruction results using Real X-Ray framework.

Table 1 reports quantitative CT synthesis quality (PSNR, SSIM) across axial, coronal, and sagittal views at each stage of the pipeline, compared against IRR3CT [5].The results show consistent improvement across stages and passes for PSNR in all three views, and for SSIM in the coronal and sagittal views. In the axial view, SSIM dips slightly after the second pass (Op-CycleGAN refinement) relative to the first-stage output, before partially recovering in the third pass with PTL— a pattern we attribute to the second-pass Op-CycleGAN prioritizing high-frequency structural detail over strict axial-plane similarity, which the PTL pass then partially corrects. Overall, the final (third-pass) output achieves the best PSNR across all three views and the best SSIM in the coronal and sagittal views, with an improvement of up to 14% in PSNR and 7.6% in SSIM (sagittal view) over IRR3CT [5].

### *D. CT Synthesis Results Over Real CXRs*

The main objective of this study is to transform real CXRs into high-fidelity volumetric CT views. Due to the lack of CXR-CT ground truth pairs, this section presents qualitative results only. Figure 8 illustrates sample CT slice synthesis results from real CXR images, where the first column presents the input CXR image, the second column shows the DRR image synthesized in Stage 1, and the third through fifth columns show the final axial, coronal, and sagittal CT synthesis results over the synthetic DRR images. Additional results are available on the study's GitHub page, [28].

The sample results demonstrate a similar level of performance to the DRR-based evaluation in Section III.C, with the proposed approach generating detailed, consistent anatomical structures and patterns directly from real CXR images. This indicates that the Stage 1 CXR-to-DRR transformation generalizes well to real clinical inputs, allowing the Stage 2 reconstruction network — trained only on synthetic DRR–CT pairs — to perform comparably on real CXR-derived DRRs.

### *E. Computational Complexity Analysis*

Table 2 summarizes the network size, parameter count, and inference time of each component of the proposed framework in a single CPU processing. All experiments were conducted on a workstation with a 2.2 GHz Intel Core i7 processor, 16 GB RAM, and an NVIDIA GeForce RTX 3080 GPU. The DRR-to-CT transformation network (first pass) contains 9.2M trainable parameters and requires approximately 50 ms for inference. Each multi-view Op-CycleGAN refinement generator contains 34.1M parameters and requires approximately 250 msec per inference; since each of the three anatomical views (axial, coronal, sagittal) is refined twice — once in the second pass and again in the third pass with PTL — this contributes 6 × 250 msec = 1,500 msec to the pipeline. Each CXR-to-DRR translation generator (Stage 1) contains only 2.1M parameters and requires approximately 10 msec for inference, applied to both the coronal and sagittal views (2 × 10 = 20 msec). Consequently, the complete inference pipeline — CXR-to-DRR translation, the first-pass DRR-to-CT baseline transformation, and the second- and third-pass Op-CycleGAN refinement across all three views — requires approximately 50 + 1,500 + 20 = 1,570 msec to synthesize a refined volumetric CT from dual-view chest X-ray images. Despite its multi-stage, multi-pass architecture, the proposed framework achieves inference in under two seconds, supporting its suitability for near real-time clinical applications.

**Table 2** Computational Complexity Analysis of the Proposed Framework

| Model Name | Purpose | Parameter (M) | Inference Time (msec) |
|---|---|---|---|
| DRR to CT Transformation Network | DRR to Low-Fidelity CT Transformation | 9.176 | 50 |
| Op-CycleGAN Generator | Sagittal Axis Refinement | 34.124 | 250 |
| Op-CycleGAN Generator | Axial Axis Refinement | 34.124 | 250 |
| Op-CycleGAN Generator | Coronal Axis Refinement | 34.124 | 250 |
| Op-CycleGAN Generator Xray to DRR | Sagittal Axis DRR to CXR Transformation | 2.144 | 10 |
| Op-CycleGAN Generator Xray to DRR | Coronal Axis DRR to CXR Transformation | 2.144 | 10 |

## IV. CONCLUSION

This study addresses the fundamentally challenging problem of synthesizing high-fidelity volumetric 3D CT views directly from CXRs. To address the inherent ambiguity of 2D-to-3D reconstruction and the scarcity of paired clinical CXR/CT datasets, we proposed Multi-Pass, Multi-View Blended Learning, a two-stage framework in which the second stage applies three successive learning passes combining supervised (paired) and unsupervised (unpaired) learning. Prior methods exhibit poor synthesis performance even on internal DRRs extracted from the target CT, since they rely on a single model trained in a single pass over DRR–CT pairs; regardless of model depth or complexity, this single-pass approach cannot resolve the multiple domain translations the problem requires. We instead adopted a "divide-and-conquer" strategy, decomposing the problem into a first stage of unsupervised CXR-to-DRR domain adaptation, followed by a followed by a second stage comprising a 3-pass process, namely, a supervised DRR-to-CT synthesis, an unsupervised multi-view CT slice refinement, and a final Progressive Transfer Learning (PTL) pass. This two-stage, multi-pass formulation reduces the domain discrepancy between synthetic DRRs and real clinical radiographs while improving the anatomical fidelity of the reconstructed CT volumes.

Experimental results confirm that each pass contributes to overall synthesis performance, with the exception of a minor SSIM reduction in the axial view following the second pass, discussed in Section III.C. After Stage 1 completes the unsupervised CXR-to-DRR domain translation, Stage 2 begins with a supervised first pass that captures the main anatomical structure from the multi-view synthetic DRR images. The subsequent unsupervised Op-CycleGAN passes substantially enhance local anatomical detail and improve structural consistency across axial, coronal, and sagittal views. The third and final PTL pass continues refining the reconstructed volumes beyond conventional convergence, helping the model escape local performance plateaus and produce sharper, more anatomically realistic reconstructions. Qualitative and quantitative evaluations on the LIDC-IDRI dataset show that the proposed framework outperforms recent prior methods, and qualitative results on real CXRs from CheXpert further demonstrate that the framework generalizes to real clinical inputs, producing high-fidelity volumetric CT views.

Future work will focus on employing uncertainty-aware reconstruction and anomaly detection into the framework to improve robustness, reliability and clinical safety. We also plan to conduct quantitative clinical validation using expert radiologist evaluation and larger, multi-source datasets to assess generalization, since the current quantitative evaluation is limited to LIDC-IDRI and the real-CXR evaluation remains qualitative. These directions aim to move CXR-to-CT reconstruction closer to practical clinical use, offering a more accessible and cost-effective pathway to volumetric imaging while supporting the detection of clinically relevant abnormalities.

# APPENDIX

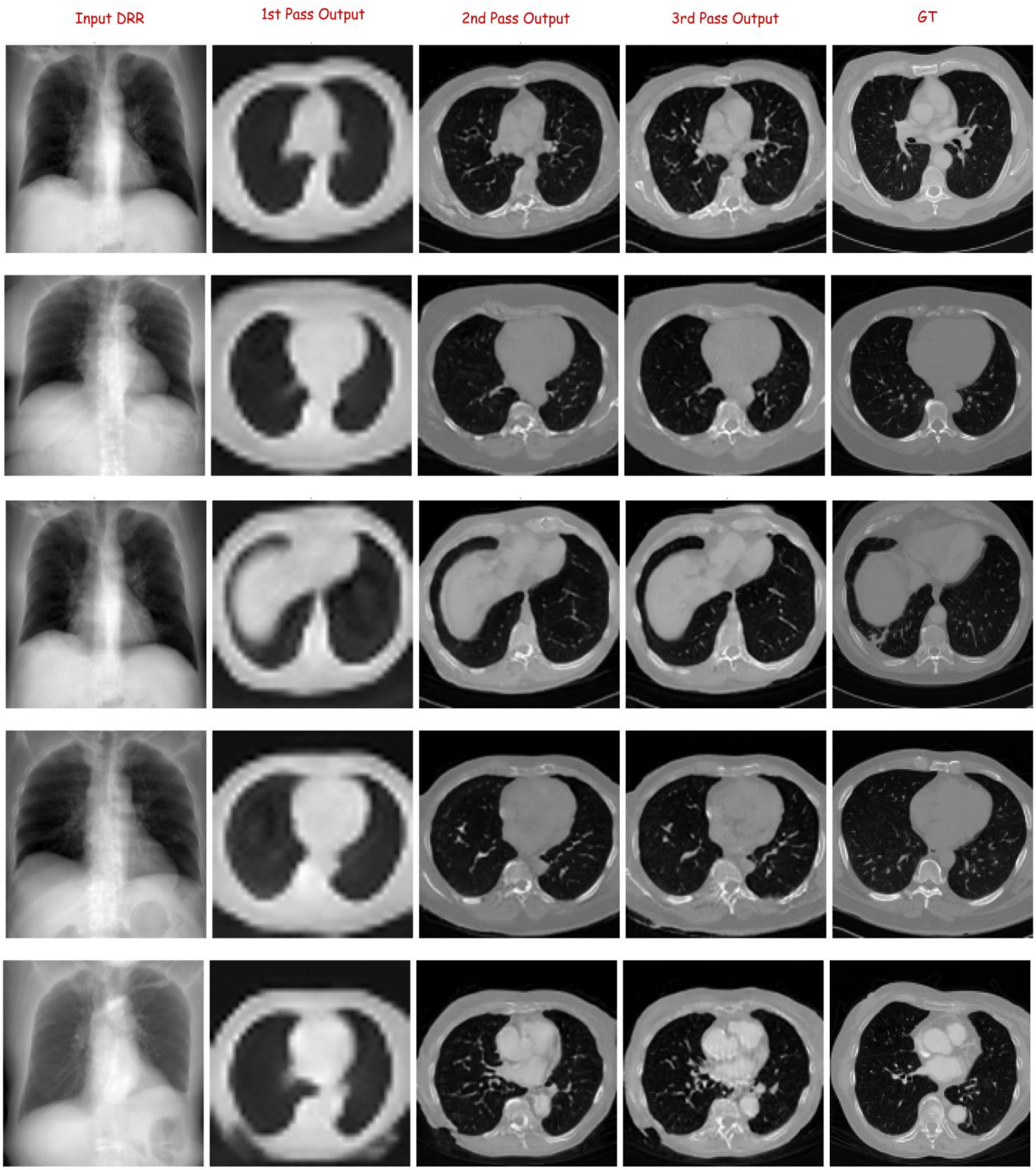

**Figure 9** Sample CT slice synthesis results at different passes of the proposed approach**.**

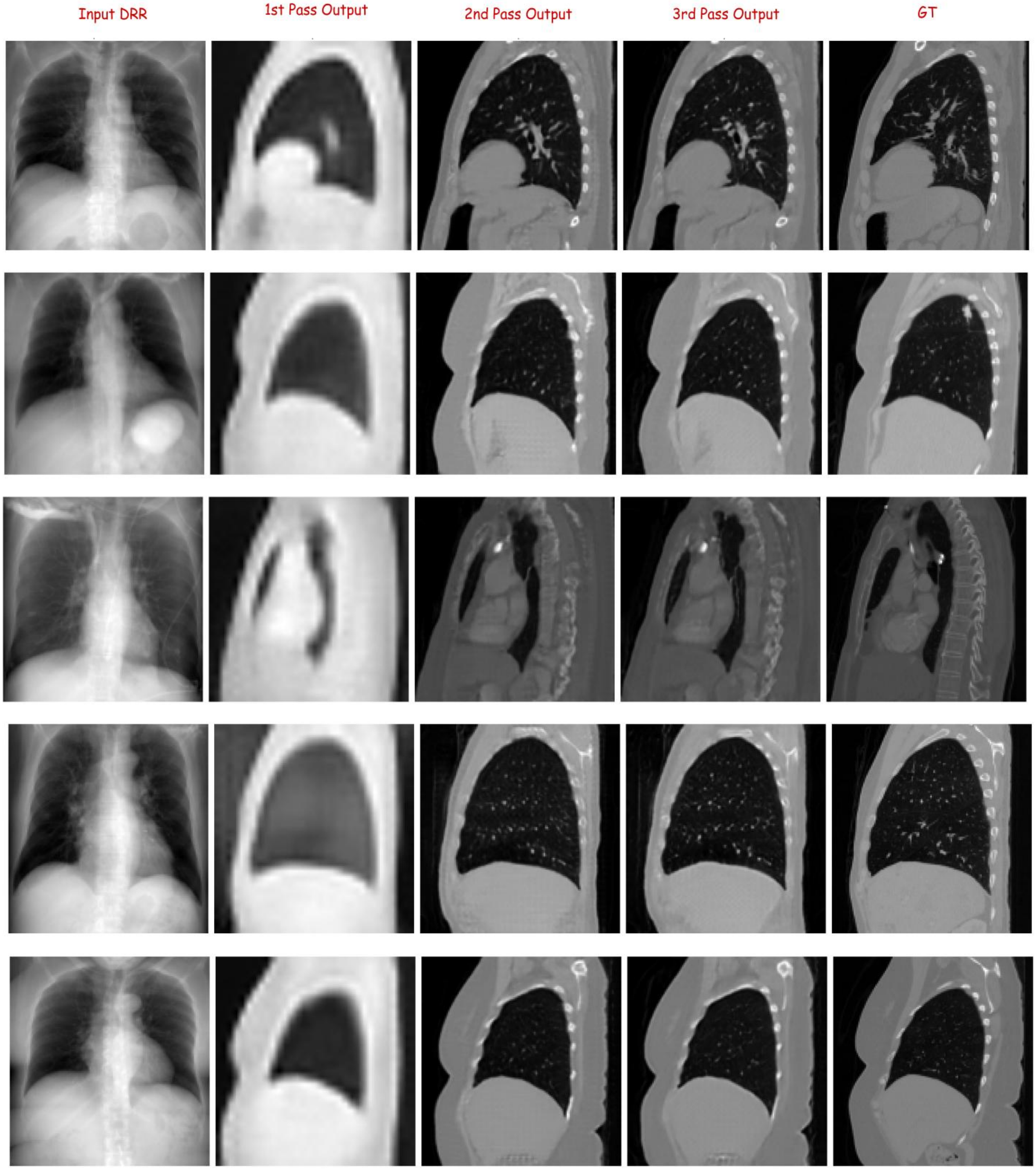

**Figure 10** Sample CT slice synthesis results at different passes of the proposed approach.

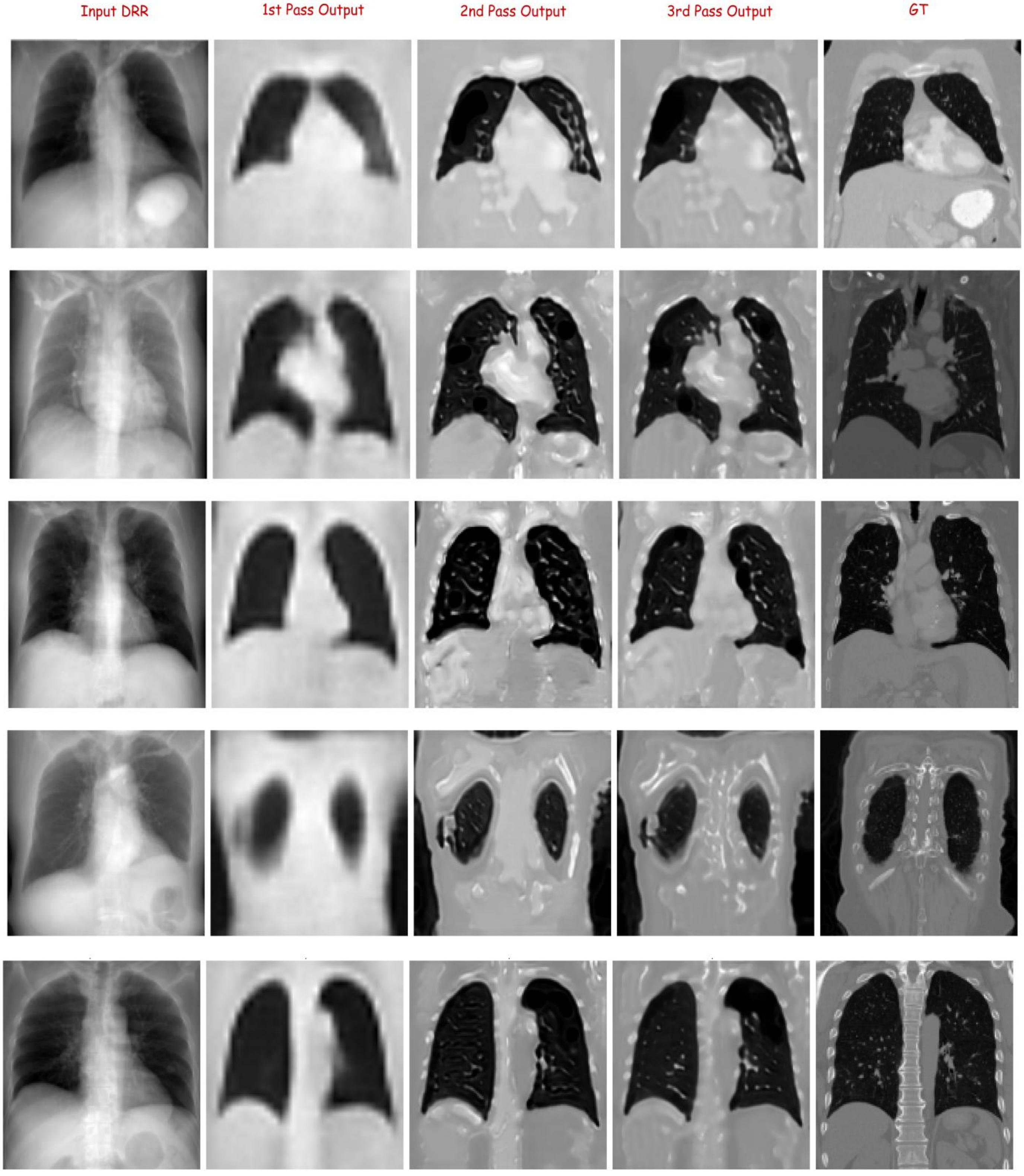

**Figure 11** Sample CT slice synthesis results at different passes of the proposed approach**.**